\pdfoutput=1

\documentclass[11pt]{article}
\usepackage[preprint]{acl}

\usepackage{times}
\usepackage{dblfloatfix}
\usepackage{latexsym}
\usepackage{amsmath} 
\usepackage{booktabs}
\usepackage{threeparttable}

\usepackage{algpseudocode}
\usepackage{algorithm}
\usepackage{multirow} 
\usepackage{fix-cm}
\usepackage{enumitem}

\usepackage[T1]{fontenc}

\usepackage[utf8]{inputenc}

\usepackage{microtype}

\usepackage{inconsolata}

\usepackage{graphicx}
\usepackage{amsfonts}

\usepackage{float}

\usepackage{cuted}

\usepackage{rotating}
\usepackage{tablefootnote}
\usepackage{url}
\usepackage[capitalize,noabbrev,nameinlink]{cleveref}

\title{{\color{blue} SELF}: {\color{blue} S}elf-{\color{blue} E}xtend the Context Length With \\ {\color{blue} L}ogistic Growth {\color{blue} F}unction}

\author{Phat Thanh Dang\thanks{Equal contributions.}$^{1}$~~Saahil Thoppay$^{*}$$^{1}$~~Wang Yang$^{1}$~~Qifan Wang$^{2}$\\\textbf{Vipin Chaudhary$^{1}$~~Xiaotian Han$^{1}$}\\
  $^{1}$Case Western Reserve University~~$^{2}$Meta \\
  \texttt{\{ptd18,svt21,wxy320,vxc204,xhan\}@case.edu~~wqfcr@meta.com} \\
}

\begin{document}
\maketitle

\begin{abstract}

Large language models suffer issues when operated on long contexts that are larger than their training context length due to the standard position encoding for tokens in the attention layer. Tokens a long distance apart will rarely have an effect on each other and long prompts yield unexpected results. To solve this problem, we propose \textit{SELF (Self-Extend the Context Length With Logistic Growth Function)}: a solution of grouping consecutive tokens at varying group sizes using a logistic capacity equation combined with a constant group size at smaller relative distances. Our model had an increase in performance of up to $12\%$ compared to the LongLM extension method in LEval (specifically on the Qwen model). On summarization related tasks in LongBench, our model performed up to $6.4\%$ better than LongLM (specifically on the Llama-2-7b model). On reading comprehension tasks from LEval, our model performed up to $5.4\%$ better than the LongLM. Our code is available at \url{https://github.com/alexeipc/SELF-LLM}.

\end{abstract}

\section{Introduction}

Large language models (LLMs) are typically pretrained on sequences with fixed maximum context lengths (e.g., 2k–4k tokens), limiting their ability to reason over or generate responses based on longer inputs. When the context length exceeds the pretraining context length, the output is severely degraded and can become unreadable and undecipherable \cite{xiao2024efficientstreaminglanguagemodels, peng2023yarnefficientcontextwindow, han2024lminfinitezeroshotextremelength, 
 chen2023extendingcontextwindowlarge,xiong2023effectivelongcontextscalingfoundation}. The main reason why the output is unpredictable when dealing with long context is Out-of-distribution (O.O.D) issues of the relative positional for LLMs using RoPE \cite{liu2023outofdistributiongeneralizationsurvey,bai2021recentadvancesadversarialtraining,zhang2022neuralnetworksfailgeneralize}. When encountering relative distances on which models were not trained, model seems to generate unpredictable output vectors that cannot be decoded by the tokenizer.

\begin{figure}[t]
    \centering
    \includegraphics[trim={1cm 4.5cm 2cm 2cm},clip,width=1\linewidth]{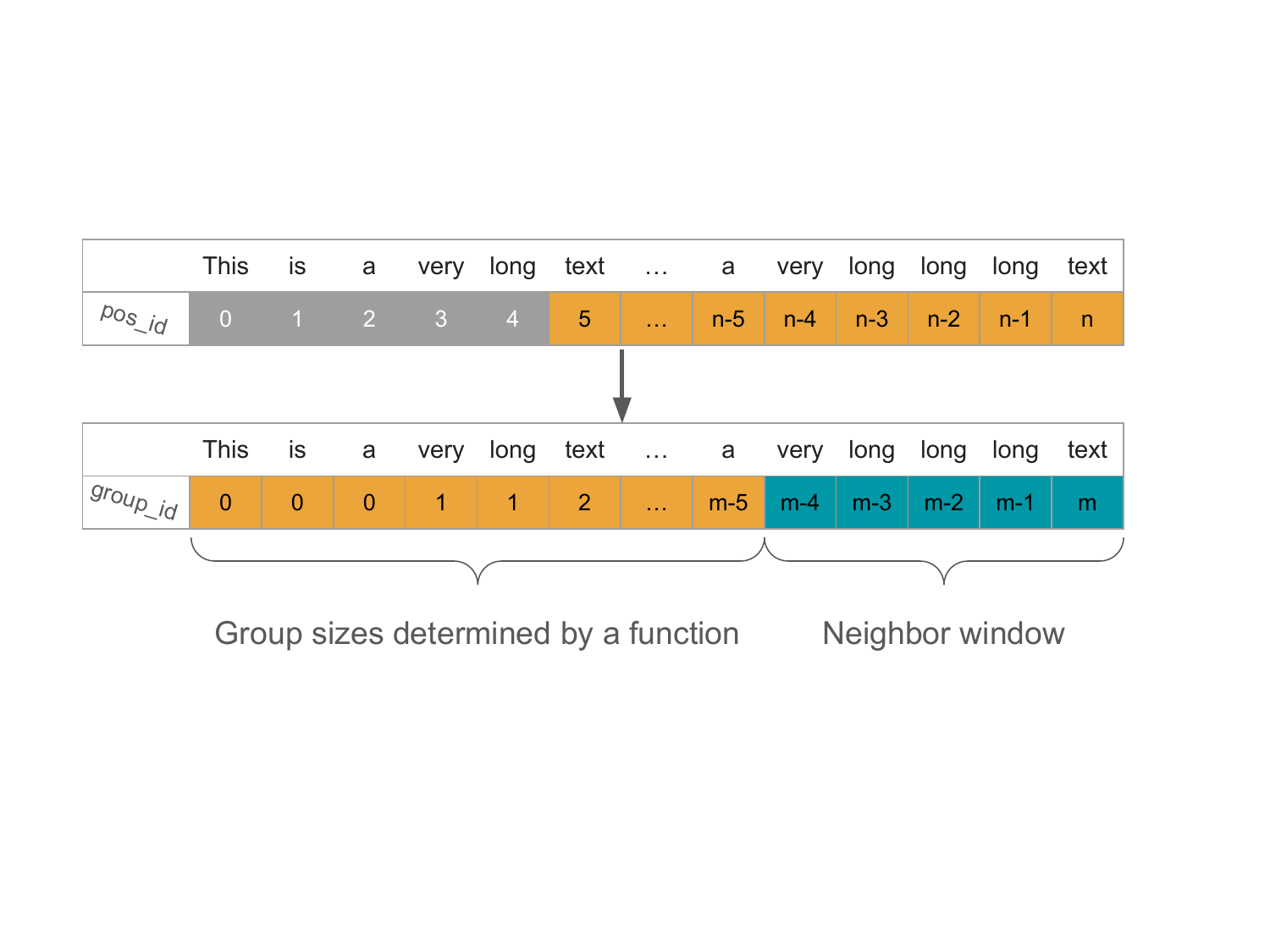}\vspace{-18pt}
    \caption{
    Illustration of our method in extending context length. Given a sequence of length $n$, that is larger than the training sequence length, the model groups consecutive tokens into groups whose sizes are determined by a function with the help of the neighbor window. As a result, the greatest index is now $m < n$, and the sequence now can be fully in the model's scope.
    }
    \label{fig:pipeline}
    \vspace{-10pt}
\end{figure}

The most intuitive way is to fine-tune the models to extend the context windows, which needs high-quality long-context data and comes with a trade-off in the performance of short-context tasks~\citep{chen2023extendingcontextwindowlarge, chen2023extendingcontextwindowlarge, zhu2024poseefficientcontextwindow}. Thus, there exist some training-free methods. For example,\citet{jin2024llmmaybelonglmselfextend} introduced Self-Extend, which leverages the model’s inherent ability to generalize to out-of-distribution (O.O.D) contexts by remapping untrained relative distances to those observed during training. This is done by grouping consecutive tokens into fixed-size chunks, combined with a neighbor window for nearby tokens.

While LongLM’s method shows promising results on long-context tasks, we propose a more adaptive strategy grounded in the observation that, in natural language, the relevance of a token typically decreases with its distance from the current context. This suggests that distant tokens can be grouped into larger units without significantly harming comprehension. Based on this intuition, we introduce a dynamic grouping strategy where group sizes increase with distance from the query. Unlike fixed-size chunking, our approach determines group boundaries through a distance-aware function, allowing for more efficient use of context length while preserving semantic fidelity.

Thus, we introduce \textit{SELF (Self-Extend the Context Length With Logistic Growth Function)}, a more adaptive token grouping strategy that dynamically adjusts group boundaries based on context structure rather than relying on fixed-size chunks. This allows for better capture of long-range dependencies and finer preservation of semantic boundaries with different distances. In essence, our method addresses the O.O.D. challenge through a shared principle but differs in the way token groups are constructed (see \cref{fig:pipeline} 
\footnote{This is an oversimplification of how the method works. More details will be explained in Our Proposal section} for illustration).

Through comprehensive experimental results, we witness an increase up to $8\%$ (specifically Qwen2-7B model \cite{qwen2}) of accuracy when applying our grouping method compared to Self-Extend \cite{jin2024llmmaybelonglmselfextend} when benchmarking on LEval. We also witnessed an accuracy increase of up to $5\%$  when using SELF over LongLM on the Llama-2-7B \cite{touvron2023llama}. 

\section{Motivations}

\begin{figure}[t]
    \centering
    \includegraphics[width=1\linewidth]{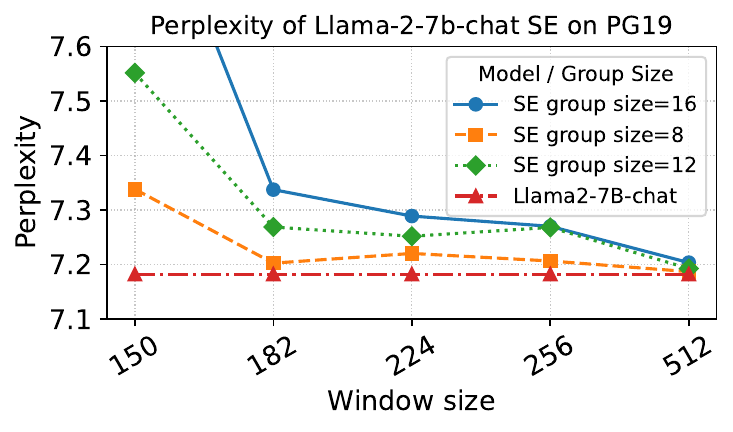}\vspace{-10pt}
    \caption{
    Illustration about relation between neighbor window and perplexity after applying Self-Extend \cite{jin2024llmmaybelonglmselfextend}. The results is derived from testing Llama-2-7B and its Self-Extend variants on the first book in PG19 \cite{raecompressive2019} with sequences of 2048 tokens. The perplexity of models applying Self-Extend slowly approaches the perplexity of the original model when increasing the neighbor window size.
    }
    \label{fig:disadvantageoflonglm}\vspace{-10pt}
\end{figure}

LongLM \cite{jin2024llmmaybelonglmselfextend} proposes a solution to handle prompts that are longer than the models' pretraining sequence lengths by grouping tokens at far distances because the exact position is less important than the relative order of information in long context and keep the exact positions of closer token by a neighbor window. However, the output's perplexity will increase right after where the neighbor window ended. As a result, to make the model less "confused", the neighbor window has to be increased (see \cref{fig:disadvantageoflonglm}), which decreases the total number of tokens the models after applying Self-Extend can handle.

Although LongLM's method yielded significantly improved results in key retrieval, the LongBench benchmark \cite{bai2024longbenchbilingualmultitaskbenchmark} and the LEval benchmark \cite{an2023levalinstitutingstandardizedevaluation}, we believe that we can increase the total context length by leveraging a property of natural language: the farther a word is from the current token, the less important it tends to be to the current context.

In LongLM \cite{jin2024llmmaybelonglmselfextend}, the group size is the same for every group, which is not the most optimized way. Intuitively, in natural languages, the further a token, the less relevant the token is to the context, allowing us to group them into progressively larger groups, meaning that we can improve the total context length by allowing larger groups without significantly trade-off in model's comprehension ability. By this intuition, the group sizes have to be dynamic, which means that they have to be determined by a function.

Therefore, we need to choose a monotonic increasing function for group sizes (the further, the larger the group). However, if the group size is too large, it will affect the performance because every word is treated the same regardless of their positions. Therefore, we must choose a function whose maximum value is limited and controllable.

Based on those conditions, we decided to choose the Logistic growth Function, which is a monotonic increasing function with a defined maximum value. Because the group sizes have to be integers, we will take the floor of the Logistic growth Function. 

$$f(x)=\lfloor \frac{Ce^{rx}}{C+e^{rx}-1} \rfloor,$$
$C$ is the capacity i.e. the maximum group size and $r$ is the growth rate of group sizes.

\section{Preliminaries}
\subsection{Position Encoding}
Most models use two types of position encodings, relative and absolute position encoding. Relative position encoding utilizes the distance between one token and another token while absolute position encoding uses the token's position from 0. \cite{vaswani2023attentionneed} Since the importance of words is usually based on how far they are away from the base word, relative position encoding is more common. Examples of absolute encoding include GPT3, learned positional encoding \cite{brown2020languagemodelsfewshotlearners}, OPT \cite{zhang2022optopenpretrainedtransformer}. Examples of relative encoding include T5, learnable attention bias \cite{xue2021mt5massivelymultilingualpretrained}, Transformer-XL \cite{dai2019transformerxlattentivelanguagemodels}, Alibi, fixed linear attention \cite{press2022trainshorttestlong}. This is especially important when it comes to long context prompts as our LLM might need to consider tokens further away as still being important. These position encodings are applied at the attention layer so that when tokens are interconnected with eachother the positions are considered. Our goal is to design a mechanism where we can consider tokens far apart in our decision making while also holding closer tokens to a high importance. Considering an example of long context key retrieval, we need to consider the close-by tokens (instructions) to a high degree but also ensure the key (at a far away position) is considered.

\subsection{RoPE}
Considering tokens $a_1\dots a_n$ and their embeddings $x_1\dots x_n$ where each embedding is a real matrix. RoPE \cite{su2023roformerenhancedtransformerrotary} integrates the position information into the query and key vectors. If done properly, $q^Tk$ will already contain the positional embeddings preventing an extra step from being needed. To embed the position, RoPE uses the function $q_m=f_q(x_m,m)\in\mathbb{R}^{|n|},k_n=f_x(x_n,n)\in\mathbb{R}^{|n|}$ where $|L|$ is the hidden dimension of each head. $f_q(x_m,m)=W_qx_me^{im\theta},f_k(x_n,n)=W_kx_ne^{in\theta},\theta_d=b^{-2d/|D|}$. The positional embedding system keeps the real section of $q^Tk$ which is $\text{Re}(q^*k)$. The dot product of the query and key will always yield a result depending on the relative distance between the two tokens as follows:
\begin{align}\label{ec:rope_equation}
&\langle f_q(x_m,m), f_k(x_n,n)\rangle_\mathbb{R} \\
&= \text{Re}(\langle f_q(x_m,m), f_k(x_n,n)\rangle_\mathbb{C}) \notag \\
&= \text{Re}(x_m^*W_q^*W_k^*x_n e^{i\theta(m-n)}) \notag \\
&= g(x_m,x_n,m-n), \notag
\end{align}
where $g$ is an abstract mapping function.

\section{Our Proposal}

\subsection{Self-Extend with constant group size}

Self-Extend\cite{jin2024llmmaybelonglmselfextend} maps unseen relative positions to trained relative positions by using the \textsc{FLOOR} operation to group neighboring tokens into one single group that shares the same positional index. 

Their important finding is the importance of neighbor attention. By just purely grouping tokens together, the perplexity will be higher than in the original model. Grouping all tokens with a constant group size degrades the effect of closer tokens which usually have more importance. To solve this problem, LongLM uses separate grouped attention, for tokens further away, and neighbor attention, for nearby tokens. Acknowledging this, our method will also apply neighbor attention.

\subsection{SELF: Self-Extend with dynamic group size}

Despite successfully tricking the LLM into believing that the tokens are closer than they really are, LongLM's approach abruptly increases the group size from $1$ (within the neighbor window, the group size can be regarded as $1$) to a much larger value (the value of group size, e.g., $512$). To avoid this sudden jump, we propose that group sizes should increase gradually rather than all at once. More specifically, the group size should grow according to a smooth function such as the Logistic Growth Function, which starts small and increases steadily. Based on this idea, we propose a new method called
\textit{SELF (Self-Extend the Context Length With Logistic Growth Function)}.

In SELF, we use a function $f:\mathbb{N} \rightarrow \mathbb{N}$ to determine the size of each group. Given a group index (like the 0th, 1st group), this function returns the number of tokens assigned to that group.

\textbf{Example 1}\label{ex:grouping}:
Given a function $f$ whose $f(0)=1,f(1)=2,f(2)=2, f(3)=3$ and $f(4)=3$. The grouping will be:
\[
F = [0,1,1,2,2,3,3,3,4,4,4]
\]

Let’s define:
\begin{itemize}[leftmargin=0.4cm, itemindent=.0cm, itemsep=0.0cm, topsep=0.1cm]
    \item $G^K:\mathbb{N}\rightarrow\mathbb{N}$ as the group position index used in the encoding of the key-value pairs.
\begin{equation}\label{ec:find_group_key_id}
G_i^K=F_i
\end{equation}

\item $G^Q:\mathbb{N}\rightarrow\mathbb{N}$ as the group position index used in the encoding of the query.

\item $R:\mathbb{N}\times\mathbb{N}\rightarrow\mathbb{N}$ is the relative distance between $G^Q$ and $G^K$.

\end{itemize}

\begin{figure}[t]
    \centering
    \includegraphics[width=1\linewidth]{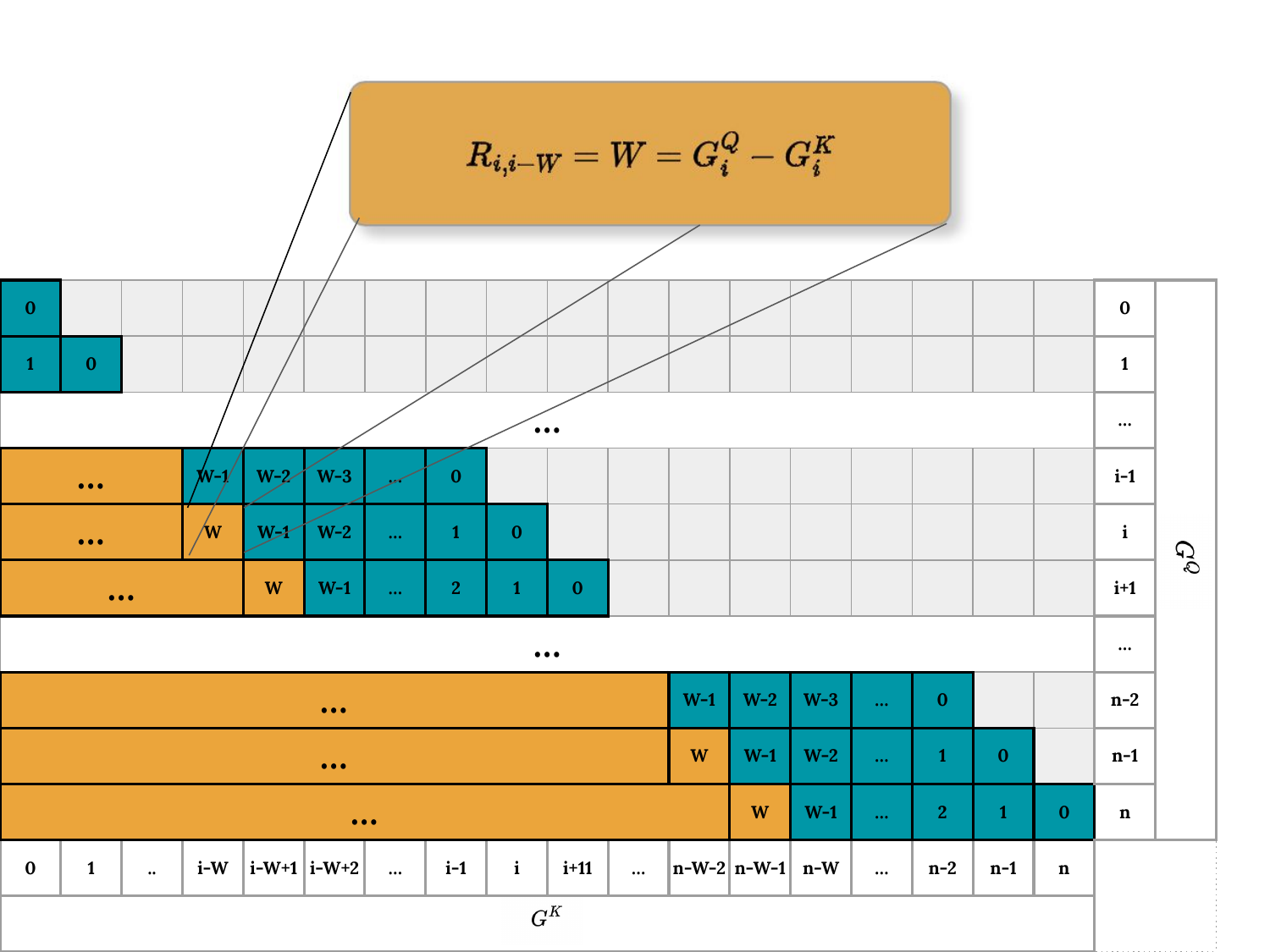}\vspace{-5pt}
    \caption{
    Illustration of the relation between $G^K$ and $G^Q$ knowing that the relative position right after the neighbor window has to be $W$.
    }
    \label{fig:enter-label}
\end{figure}

The relative position right after the neighbor window ($R_{i,i-W}$) has to be $W$, where $W$ is the width of the neighbor window, because the relative positions inside the neighbor window will range from $0$ to $W-1$ (see illustration in \cref{fig:enter-label}). Therefore, 
\begin{equation}\label{ec:find_group_query_id}
    G^Q_i=
\begin{cases}
                W+G^K_{i-W}, & \text{if } i \leq W \\
                c, & \text{otherwise}
\end{cases}
\end{equation}

\begin{figure*}[t]
    
\includegraphics[trim={0cm 5.5cm 0cm 4.5cm},clip,width=1\textwidth]{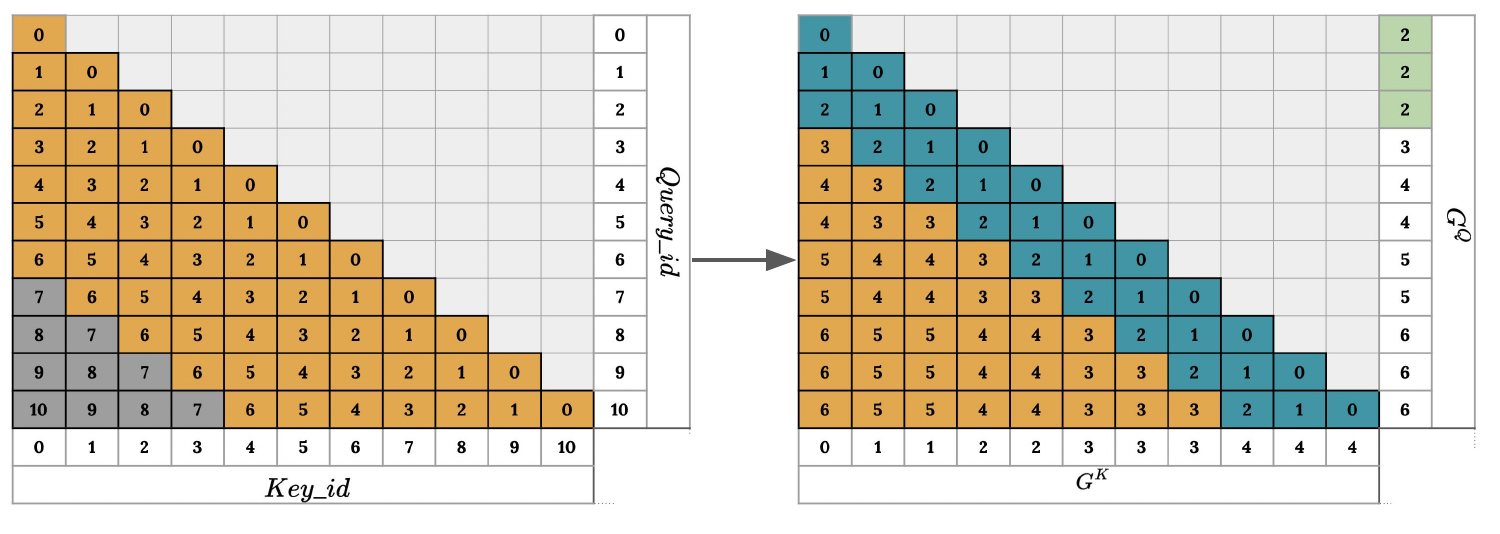}\vspace{-15pt}
    \caption{ Illustration of the algorithm grouping the indices using the function $f:\mathbb{N}\rightarrow \mathbb{N}$, where $f(0)=1,f(1)=2,f(2)=2, f(3)=3$ and $f(4)=3$. The sequence with length of $n=11$ was run the model with the pretraining sequence length of $L=6$. The numbers denote the relative position between the corresponding key and query token. It has two kinds of self-attention, similar to Self-Extend \cite{jin2024llmmaybelonglmselfextend}: neighbor tokens inside the neighbor window ($W=3$) (blue cells in the figure) use regular self-attention; group tokens outside the neighbor window (orange cells in the figure) use group self-attention (group indices are denoted as the $G$ row and column in the figure). Green $G^Q$ means it can be anything as it is covered completed by the neighbor window.}
    \label{fig:group-query-id}
    \vspace{-10pt}
\end{figure*}

No matter what constant $c$ we choose, it will be completely covered by the neighbor window.

If we used only group attention, the number of tokens can be fully extended to $\sum_{i=1}^{L}{f(i)}$. However, because we have the neighbor window, $R_{n,n-W}=W$ instead of $R_{n,n-W}=\max(F)-F_W$, that is, it takes $W+F_W-\max(F)$ more indices than using only group attention. Therefore, the number of tokens can be extended to 
$$
    L'=\sum_{i=1}^{L+\max(F)-W-F_W}{f(i)}
$$  
where $L$ is the initial token limit.

This formula gives the total number of tokens that can be processed using our SELF method, which blends regular and group-based attention in a way that grows group sizes smoothly and avoids any sudden jumps that could disrupt the model.

\subsection{Efficient Implementation: grouping indices in parallel}
The most naive approach to calculate $F$ given $f$ is to start with an empty $F$, than sequentially compute $f(i)$ and add $f(i)$ more elements to the end of $F$.
$$F \gets F \mathbin{\|} \operatorname{replicate}(i, f(i))$$
However, since we have thousands of tokens, computing the new positional embeddings sequentially would take $O(n)$ of run time.

In order to solve this high computing time issue, we use inverse function of the grouping function which will divide each sequence into sections in which the group sizes are the same so we can easily calculate and assign group sizes in parallel. We define the inverse function $f^{-1}:\mathbb{N}\rightarrow \mathbb{N}$, when given the group size, the inverse function returns the smallest index that has the given group size.

Using \hyperref[ex:grouping]{Example 1}, the inverse function will be $f^{-1}(1)=0, f^{-1}(2)=1$ and $f^{-1}(3)=3$. Let's define the function $g:\mathbb{N}\rightarrow \mathbb{N}$, when given the group size, the function returns the total number of elements that are in the group of the given size.
\begin{equation}\label{ec:calculate_g}
    g(x)=x\cdot [f^{-1}(x+1)-f^{-1}(x) + 1]
\end{equation}

In order to find the $F_i$ given $i$, define $k$ the largest number such that:
\begin{equation}\label{ec:find_k_equation}
    S=\sum_{j=1}^k g(j)<i
\end{equation}

This means that $F_i$ has to be in the group whose size is $k+1$ and it is $\lceil\frac{i-S}{k+1}\rceil$ indices away from the last index of the group whose size is $k$, which is $f^{-1}(k+1)-1$. Therefore, 
\begin{equation}\label{ec:find_F_i_equation}
F_i=[f^{-1}(k+1) - 1] +\lceil\frac{i-S}{k+1}\rceil
\end{equation}

\begin{algorithm}[t]\label{code:pseudo-code}
    \small
    \caption{Construct group indices($n,W,C,r$)}
    \begin{algorithmic}
        \State $p\gets -1$
        \State
        \For {$k\gets 1$ to $C-1$}
            \State Compute parallelly $F[p+1..p+g(k)]$ using \cref{ec:find_F_i_equation}
            \State $p\gets p+g(k)$
        \EndFor
        \State
        \State Compute $G^K$ parallelly using \cref{ec:find_group_key_id}
        \State Compute $G^Q$ parallelly using \cref{ec:find_group_query_id}
        \State
        \State \Return $group\_key\_id$ and $group\_query\_id$ 
    \end{algorithmic}
\end{algorithm}

Now, considering the logistic growth function, we have:
$$f(x)=\lfloor \frac{Ce^{rx}}{C+e^{rx}-1} \rfloor$$
$$f^{-1}(y)=\lfloor\frac{\ln(Cy-y)-\ln(C-y)}{r}\rfloor$$

In the logistic growth function, the maximum group size $k$ in \cref{ec:find_k_equation} is $C$, which is a very small number compared to the sequence length. We can utilize GPU parallelism using the \hyperref[code:pseudo-code]{pseudo-code}. By applying \cref{ec:calculate_g}, $g(k)$ can be computed in $O(1)$. We can easily tell that the total work for computing $F$ is $O(n+C)$, and the total work for computing $G^K$ and $G^Q$ knowing $F$ is $O(n)$, since it takes $O(1)$ at each index.

\begin{table*}[t]
\centering
\caption{Perplexity on dataset PG19 \cite{raecompressive2019} first book with Llama-2-7b-chat and compare SE and SELF (the growth rate $r=0.02$) with the same group size ($C=16$, $C=32$ and $C=64$) and neighbor window.}\label{table:ppl}
\vspace{-10pt}
\setlength{\tabcolsep}{8pt}
    \begin{tabular}{@{}llcccccccc@{}}
    \toprule
    {}& Model & \multicolumn{6}{c}{Perplexity with Context Window Size (log scale)} \\ 
    {}&\textbf{Name}   & \textbf{4096} &\textbf{6144} & \textbf{8192} & \textbf{10240} & \textbf{12288} &  \textbf{14336} & \textbf{16384} \\
    \midrule
    {}&Llama-2-7b-chat &7.231 &$> 10^3$ &$> 10^3$ &$> 10^3$ &$> 10^3$ &$> 10^3$ &$> 10^3$  \\
    \midrule
    \multirow{2}{*}{\rotatebox[origin=c]{90}{\fontsize{5}{100} $C=16$}}&SE-Llama-2-7b-chat  &7.103 &7.086 &7.126 &7.174 &7.229 &7.248 &7.270 \\
    {}&SELF-Llama-2-7b-chat &7.085 &7.085 &7.122 &7.168 &7.203 &7.234 &7.270    \\
    \midrule
    \multirow{2}{*}{\rotatebox[origin=c]{90}{\fontsize{5}{100} $C=32$}}&SE-Llama-2-7b-chat  &7.141 &7.184 &7.199 &7.314 &7.346 &7.410 &7.496 \\
    {}&SELF-Llama-2-7b-chat  &7.119 &7.133 &7.196 &7.275 &7.345 &7.408 &7.484 \\
    \midrule
    \multirow{2}{*}{\rotatebox[origin=c]{90}{\fontsize{5}{100} $C=64$}}&SE-Llama-2-7b-chat &7.186 &7.316 &7.303 &7.458 &7.530 &7.625 &8.041  \\
    {}&SELF-Llama-2-7b-chat  &7.135 &7.180 &7.267 &7.364 &7.467 &7.619 &8.068  \\

    \bottomrule
    \end{tabular}
\end{table*}

Putting this together, the total work for the algorithm is $O(n+C)$ and the parallel span is $O(C)$. Assuming that there are $P$ threads available, the runtime is bounded by:
$$T(P)=O(\max\{\frac{n+C}{P},C\})$$

This means that if having sufficient resources ($P$ is large enough), we can speed up to near-linear since the lower bound is $C$, which is usually a very small number.

\section{Experiments}

In this section, we first analyze the impact of \textit{Group Size} on the \textsc{SELF} method based on perplexity results from the PG19 dataset, in order to identify an appropriate group size configuration. We then compare \textsc{SELF} with the standard \textsc{SE} method on real-world long-context benchmarks such as \textsc{LongBench} and \textsc{LEval}, demonstrating the effectiveness of \textsc{SELF} on practical long-conetxt tasks.

We ran experiments on Llama-2-7B, Llama-2-13B \cite{touvron2023llama2openfoundation}, Qwen-7b \cite{qwen2}, and a distilled reasoning model from Deepseek-R1 \cite{deepseekai2025deepseekr1incentivizingreasoningcapability}. We conducted our tests on Longbench v1 \cite{bai2024longbenchbilingualmultitaskbenchmark} and LEval \cite{an2023levalinstitutingstandardizedevaluation}\footnote{We skipped the GSM benchmark as we were unable to replicate the results the paper provided on our own}.

\subsection{Understanding the impact of group size on SELF via Perplexity}

We begin by measuring the perplexity of LLaMA-2-7B-Chat with both \textsc{SE} and \textsc{SELF} under varying group sizes and context window lengths on the PG19 dataset. From \cref{table:ppl}, we can observe:
\begin{itemize}[leftmargin=0.4cm, itemindent=.0cm, itemsep=0.0cm, topsep=0.1cm]
    \item SELF achieves lower perplexity scores when working with the same group size.
    \item When group size is small, there isn't much difference in perplexity score between SE and SELF.
    \item The larger the group size, the longer the sequence lengths over which noticeable differences can be observed. For example, the difference is still noticeable when $C=32$, and the sequence length is $6144$ and when $C=64$ and  the sequence length is $12288$. With larger group size, it takes longer sequence to reach its maximum group size, meaning that most of the groups has their sizes less than the maximum group size, decreasing the final perplexity score.
    \item When dealing with sequences that are significantly longer than the original context length, the scores are basically the same for SE and SELF. When the sequence is significantly long, the amount of intermediate group is negligible compared to the number of groups that have reached the maximum group size. As a result, the model behavior closely resembles that of SE, where all groups have the maximum group size.
\end{itemize}

From the above observations, there is a trade off in perplexity when increasing the group size \cite{jin2024llmmaybelonglmselfextend}. When using larger group size, models are more uncertain in their predictions. However, when dealing with the same group size\footnote{Group size in context of SELF refer to the maximum group size i.e. the capacity in the Logistic growth function}, ideally, models with SELF have lower perplexity than ones with SE because instead of increasing rapidly from one to the maximum group size, models with SELF have to go through smaller group sizes (intermediate groups) before reaching the maximum group size. Therefore, although the context length after extension of both methods are approximately the same with the same group size, SELF performs better when the sequence length is not excessively larger than the orginial context window.

Base on this analysis, we decided to choose larger group size for most of our experiments compared to ones in LongLM \cite{jin2024llmmaybelonglmselfextend} without experiencing a significant trade off in performance.

\subsection{Comparisons of SELF and SE in Real-World Long Context Tasks}
\subsubsection{LongBench}
\begin{table*}[t]
\fontsize{18}{24}\selectfont
\setlength{\tabcolsep}{3.5pt}
\centering
\caption{Performance of different models on LongBench \cite{bai2024longbenchbilingualmultitaskbenchmark}. {\color{blue}*} means that the results are reported by Self-Extend \cite{jin2024llmmaybelonglmselfextend},  {\color{red}*} means that the results are run by us (single run). The suffix number (e.g. ‘25k’) indicates the maximum context window of the model. The ‘SE’ prefix indicates
Self-Extend is applied to this model and the prefix ‘SEF’ indicates that our Self-Extend with Logistic Growth Function is applied. The best performance in each section will be in \textbf{bold}}\label{tab:longbench}
\vspace{-10pt}

\begin{threeparttable}

\scalebox{0.3}{
\begin{tabular}{l|lcccccccccccccccc}
\specialrule{1pt}{0pt}{2pt}
&\multirow{4}{*}{~~~~~~~~~~~~~~~~~~~~LLMs\tnote{a}} & \multicolumn{3}{c}{Single-Document QA} & \multicolumn{3}{c}{Multi-Document QA}& \multicolumn{3}{c}{Summarization}& \multicolumn{3}{c}{Few-shot Learning}& \multicolumn{2}{c}{Synthetic} & \multicolumn{2}{c}{Code} \\
\cmidrule(lr){3-5}\cmidrule(lr){6-8}\cmidrule(lr){9-11}\cmidrule(lr){12-14}\cmidrule(lr){15-16}\cmidrule(lr){17-18}
&& \rotatebox[origin=c]{30}{NarrativeQA} & \rotatebox[origin=c]{30}{Qasper} & \rotatebox[origin=c]{30}{MultiField-en} & \rotatebox[origin=c]{30}{HotpotQA} & \rotatebox[origin=c]{30}{2WikiMQA} & \rotatebox[origin=c]{30}{Musique} & \rotatebox[origin=c]{30}{GovReport} & \rotatebox[origin=c]{30}{QMSum} & \rotatebox[origin=c]{30}{MultiNews} & \rotatebox[origin=c]{30}{TREC} & \rotatebox[origin=c]{30}{TriviaQA} & \rotatebox[origin=c]{30}{SAMSum} & \rotatebox[origin=c]{30}{PassageCount} & \rotatebox[origin=c]{30}{PassageRe} & \rotatebox[origin=c]{30}{Lcc} & \rotatebox[origin=c]{30}{RepoBench-P} \\

\specialrule{1pt}{2pt}{2pt}

\multirow{4}{*}{\rotatebox[origin=c]{90}{\fontsize{18}{100}\selectfont SE vs SELF}}&\cellcolor{green!10}~~~~~~Llama-2-7B-chat-4k{\color{blue}*} & \cellcolor{green!10}18.7&\cellcolor{green!10}19.2&\cellcolor{green!10} 36.8 &\cellcolor{green!10} 25.4 &\cellcolor{green!10} 32.8&\cellcolor{green!10} 9.4 &\cellcolor{green!10} 27.3 & \cellcolor{green!10}20.8 &\cellcolor{green!10} 25.8 &\cellcolor{green!10} 61.5 &\cellcolor{green!10} 77.8 & \cellcolor{green!10} 40.7 &\cellcolor{green!10} 2.1 &\cellcolor{green!10}\textbf{9.8} & \cellcolor{green!10}52.4 & \cellcolor{green!10}43.8\\
&\cellcolor{green!10}SE-Llama-2-7B-chat-16k{\color{blue} *}&\cellcolor{green!10}\textbf{21.69}&\cellcolor{green!10}25.02&\cellcolor{green!10}35.21&\cellcolor{green!10}34.34&\cellcolor{green!10}30.24            &\cellcolor{green!10}14.13         &\cellcolor{green!10}27.32&\cellcolor{green!10}21.35&\cellcolor{green!10}25.78&\cellcolor{green!10}\textbf{69.50} &\cellcolor{green!10}\textbf{81.99}&\cellcolor{green!10}40.96&\cellcolor{green!10}5.66&\cellcolor{green!10}5.83&\cellcolor{green!10}\textbf{60.60}&\cellcolor{green!10}\textbf{54.33}\\ 
&\cellcolor{green!10}SE-Llama-2-7B-chat-25k{\color{blue} *}&\cellcolor{green!10}21.37&\cellcolor{green!10}\textbf{26.68}&\cellcolor{green!10}34.63&\cellcolor{green!10}35.47   &\cellcolor{green!10}30.46&\cellcolor{green!10}\textbf{15.51}&\cellcolor{green!10}27.51&\cellcolor{green!10}21.30&\cellcolor{green!10}25.87&\cellcolor{green!10}68.50&\cellcolor{green!10}78.79&\cellcolor{green!10}\textbf{41.29}&\cellcolor{green!10}3.90            &\cellcolor{green!10}3.50&\cellcolor{green!10}59.69            &\cellcolor{green!10}53.83\\ 

&\cellcolor{green!10}SELF-Llama-2-7B-chat-100k{\color{red} *}&\cellcolor{green!10}17.4&\cellcolor{green!10}25.74&\cellcolor{green!10}\textbf{37.61}&\cellcolor{green!10} \textbf{36.30}   &\cellcolor{green!10}\textbf{31.37}&\cellcolor{green!10}13.11&\cellcolor{green!10}\textbf{27.9}&\cellcolor{green!10}\textbf{21.81}&\cellcolor{green!10}\textbf{27.54}&\cellcolor{green!10}\textbf{69.50}&\cellcolor{green!10}76.97&\cellcolor{green!10} 40.85 &\cellcolor{green!10}\textbf{6.16}           &\cellcolor{green!10}6.0&\cellcolor{green!10}60.49            &\cellcolor{green!10} 51.55\\

\specialrule{1pt}{2pt}{10pt}\specialrule{1pt}{2pt}{2pt}

\multirow{7}{*}{\rotatebox[origin=c]{90}{\fontsize{18}{100}\selectfont Other Methods}} & \cellcolor{green!10}LongChat1.5-7B-32k{\color{blue}*} & \cellcolor{green!10}16.9 & \cellcolor{green!10}27.7 & \cellcolor{green!10}41.4 & \cellcolor{green!10}31.5 & \cellcolor{green!10}20.6 & \cellcolor{green!10}9.7 & \cellcolor{green!10}30.8 & \cellcolor{green!10}22.7 & \cellcolor{green!10}26.4 & \cellcolor{green!10}63.5 & \cellcolor{green!10}82.3 & \cellcolor{green!10}34.2 & \cellcolor{green!10}1.0 & \cellcolor{green!10}\textbf{30.5} & \cellcolor{green!10}53.0 & \cellcolor{green!10}55.3 \\
& \cellcolor{green!10}together/llama-2-7b-32k{\color{blue}*} & \cellcolor{green!10} 15.65 & \cellcolor{green!10}10.49 & \cellcolor{green!10}33.43 & \cellcolor{green!10}12.36 & \cellcolor{green!10}12.53 & \cellcolor{green!10}6.19 & \cellcolor{green!10}29.28 & \cellcolor{green!10} 17.18 & \cellcolor{green!10}22.12 & \cellcolor{green!10}\textbf{71.0} & \cellcolor{green!10}\textbf{87.79} & \cellcolor{green!10} \textbf{43.78} & \cellcolor{green!10}1.0 & \cellcolor{green!10} 23.0 & \cellcolor{green!10} 63.79 & \cellcolor{green!10}\textbf{61.77} \\			
& \cellcolor{green!10}CLEX-7B-16k{\color{blue}*} & \cellcolor{green!10}18.05 & \cellcolor{green!10}23.68 & \cellcolor{green!10}\textbf{44.62} & \cellcolor{green!10}28.44 & \cellcolor{green!10}19.53 & \cellcolor{green!10}9.15 & \cellcolor{green!10}\textbf{32.52} & \cellcolor{green!10}22.9 & \cellcolor{green!10}25.55 & \cellcolor{green!10}68 & \cellcolor{green!10}84.92 & \cellcolor{green!10}42.82 & \cellcolor{green!10}0 & \cellcolor{green!10}11.5 & \cellcolor{green!10}59.01 & \cellcolor{green!10}56.87 \\
& \cellcolor{green!10}CodeLLaMA-7B-16k{\color{blue}*} & \cellcolor{green!10}\textbf{22.93} & \cellcolor{green!10}\textbf{30.69} & \cellcolor{green!10}43.37 & \cellcolor{green!10}33.05 & \cellcolor{green!10}27.93 & \cellcolor{green!10}14.2 & \cellcolor{green!10}28.43 & \cellcolor{green!10}\textbf{24.18} & \cellcolor{green!10}26.84 & \cellcolor{green!10}70 & \cellcolor{green!10}84.97 & \cellcolor{green!10}43.43 & \cellcolor{green!10}2 & \cellcolor{green!10}13.5 & \cellcolor{green!10}\textbf{64.35} & \cellcolor{green!10}55.87 \\
& \cellcolor{green!10}SE-Llama-2-7B-chat-16k{\color{blue}*} & \cellcolor{green!10}21.69 & \cellcolor{green!10}25.02 & \cellcolor{green!10}35.21 & \cellcolor{green!10}34.34 & \cellcolor{green!10}30.24 & \cellcolor{green!10}14.13 & \cellcolor{green!10}27.32 & \cellcolor{green!10}21.35 & \cellcolor{green!10}25.78 & \cellcolor{green!10}69.50 & \cellcolor{green!10}81.99 & \cellcolor{green!10}40.96 & \cellcolor{green!10}5.66 & \cellcolor{green!10}5.83 & \cellcolor{green!10}60.60 & \cellcolor{green!10}54.33 \\
& \cellcolor{green!10}SE-Llama-2-7B-chat-25k{\color{blue}*} & \cellcolor{green!10}21.37 & \cellcolor{green!10}26.68 & \cellcolor{green!10}34.63 & \cellcolor{green!10}35.47 & \cellcolor{green!10}30.46 & \cellcolor{green!10}\textbf{15.51} & \cellcolor{green!10}27.51 & \cellcolor{green!10}21.30 & \cellcolor{green!10}25.87 & \cellcolor{green!10}68.50 & \cellcolor{green!10}78.79 & \cellcolor{green!10}41.29 & \cellcolor{green!10}3.90 & \cellcolor{green!10}3.50 & \cellcolor{green!10}59.69 & \cellcolor{green!10}53.83 \\
&\cellcolor{green!10}SELF-Llama-2-7B-chat-100k{\color{red} *}&\cellcolor{green!10}17.4&\cellcolor{green!10}25.74&\cellcolor{green!10}37.61&\cellcolor{green!10} \textbf{36.30}  &\cellcolor{green!10}\textbf{31.37}&\cellcolor{green!10}13.11&\cellcolor{green!10}27.9&\cellcolor{green!10}21.81&\cellcolor{green!10}\textbf{27.54}&\cellcolor{green!10}69.50&\cellcolor{green!10}76.97&\cellcolor{green!10} 40.85 &\cellcolor{green!10}\textbf{6.16}           &\cellcolor{green!10}6.0&\cellcolor{green!10}60.49            &\cellcolor{green!10} 51.55\\ 

\specialrule{1pt}{2pt}{10pt} \specialrule{1pt}{2pt}{2pt}

\multirow{7}{*}{\rotatebox[origin=c]{90}{\fontsize{18}{100}\selectfont Fixed Models}} & \cellcolor{gray!10}GPT-3.5-Turbo-16k{\color{blue}*} & \cellcolor{gray!10}23.6 & \cellcolor{gray!10}\textbf{43.3} & \cellcolor{gray!10}\textbf{52.3} & \cellcolor{gray!10}51.6 & \cellcolor{gray!10}37.7 & \cellcolor{gray!10}26.9 & \cellcolor{gray!10}29.5 & \cellcolor{gray!10}23.4 & \cellcolor{gray!10}26.7 & \cellcolor{gray!10}68.0 & \cellcolor{gray!10}\textbf{91.4} & \cellcolor{gray!10}\textbf{41.7} & \cellcolor{gray!10}\textbf{4.5} & \cellcolor{gray!10}71.0 & \cellcolor{gray!10}54.7 & \cellcolor{gray!10}53.6 \\
& \cellcolor{gray!10}XGen-7B-8k{\color{blue}*} & \cellcolor{gray!10}18 & \cellcolor{gray!10}18.1 & \cellcolor{gray!10}37.7 & \cellcolor{gray!10}29.7 & \cellcolor{gray!10}21.1 & \cellcolor{gray!10}10.3 & \cellcolor{gray!10}27.3 & \cellcolor{gray!10}20.5 & \cellcolor{gray!10}26.2 & \cellcolor{gray!10}65.5 & \cellcolor{gray!10}77.8 & \cellcolor{gray!10}25.3 & \cellcolor{gray!10}2.1 & \cellcolor{gray!10}8.5 & \cellcolor{gray!10}38.6 & \cellcolor{gray!10}38.6 \\
& \cellcolor{gray!10}InternLM-7B-8k{\color{blue}*} & \cellcolor{gray!10}12.1 & \cellcolor{gray!10}16.7 & \cellcolor{gray!10}23.4 & \cellcolor{gray!10}28.7 & \cellcolor{gray!10}22.8 & \cellcolor{gray!10}9.0 & \cellcolor{gray!10}9.7 & \cellcolor{gray!10}15.9 & \cellcolor{gray!10}22.8 & \cellcolor{gray!10}52.0 & \cellcolor{gray!10}77.8 & \cellcolor{gray!10}21.2 & \cellcolor{gray!10}3.0 & \cellcolor{gray!10}6.0 & \cellcolor{gray!10}44.1 & \cellcolor{gray!10}28.8 \\
& \cellcolor{gray!10}ChatGLM2-6B-32k{\color{blue}*} & \cellcolor{gray!10}21.1 & \cellcolor{gray!10}31.5 & \cellcolor{gray!10}46.2 & \cellcolor{gray!10}45.1 & \cellcolor{gray!10}34.0 & \cellcolor{gray!10}21.9 & \cellcolor{gray!10}32.4 & \cellcolor{gray!10}\textbf{24.0} & \cellcolor{gray!10}26.5 & \cellcolor{gray!10}62.5 & \cellcolor{gray!10}78.7 & \cellcolor{gray!10}36.3 & \cellcolor{gray!10}1.5 & \cellcolor{gray!10}77.0 & \cellcolor{gray!10}55.6 & \cellcolor{gray!10}49.9 \\
& \cellcolor{gray!10}ChatGLM3-6B-32k{\color{blue}*} & \cellcolor{gray!10}\textbf{26.0} & \cellcolor{gray!10}\textbf{43.3} & \cellcolor{gray!10}51.7 & \cellcolor{gray!10}\textbf{54.4} & \cellcolor{gray!10}\textbf{44.9} & \cellcolor{gray!10}\textbf{40.4} & \cellcolor{gray!10}\textbf{36.8} & \cellcolor{gray!10}23.9 & \cellcolor{gray!10}\textbf{27.9} & \cellcolor{gray!10}\textbf{79.0} & \cellcolor{gray!10}87.1 & \cellcolor{gray!10}38.2 & \cellcolor{gray!10}2.0 & \cellcolor{gray!10}\textbf{99.0} & \cellcolor{gray!10}\textbf{57.66} & \cellcolor{gray!10}\textbf{54.76} \\
& \cellcolor{gray!10}Baichuan-13B-4k{\color{blue}*} & \cellcolor{gray!10}0.07 & \cellcolor{gray!10}17.55 & \cellcolor{gray!10}17.28 & \cellcolor{gray!10}3.29 & \cellcolor{gray!10}15 & \cellcolor{gray!10}0.1 & \cellcolor{gray!10}6.8 & \cellcolor{gray!10}1.71 & \cellcolor{gray!10}23.1 & \cellcolor{gray!10}20.05 & \cellcolor{gray!10}20.06 & \cellcolor{gray!10}5.77 & \cellcolor{gray!10}0.06 & \cellcolor{gray!10}0.5 & \cellcolor{gray!10}47.98 & \cellcolor{gray!10}16.58 \\
& \cellcolor{gray!10}ALiBi-7B-4k{\color{blue}*} & \cellcolor{gray!10} 0.04 & \cellcolor{gray!10} 8.13 & \cellcolor{gray!10} 17.87 & \cellcolor{gray!10} 2.73 & \cellcolor{gray!10} 8 & \cellcolor{gray!10} 1.33 & \cellcolor{gray!10} 5.31 & \cellcolor{gray!10} 1.64 & \cellcolor{gray!10} 25.55 & \cellcolor{gray!10} 9.25 & \cellcolor{gray!10} 8.83 & \cellcolor{gray!10} 4.67 & \cellcolor{gray!10} 0 & \cellcolor{gray!10} 1.27 & \cellcolor{gray!10} 46.69 & \cellcolor{gray!10} 18.54 \\
\specialrule{1pt}{2pt}{0pt}

\end{tabular}
}

\end{threeparttable}

\end{table*}

We conducted experiment on LongBench \cite{bai2024longbenchbilingualmultitaskbenchmark} using Llama-2-7B and then compared our results with the original model and the model where Self-Extend is applied. Here we decided instead of using small group size of $6$ and $8$ like in LongLM \cite{jin2024llmmaybelonglmselfextend}, we used a much bigger group size ($C=32$) and still observed a better results in most tasks. The results are in \cref{tab:longbench}
\begin{figure}[t]
    \centering
    \includegraphics[width=1\linewidth]{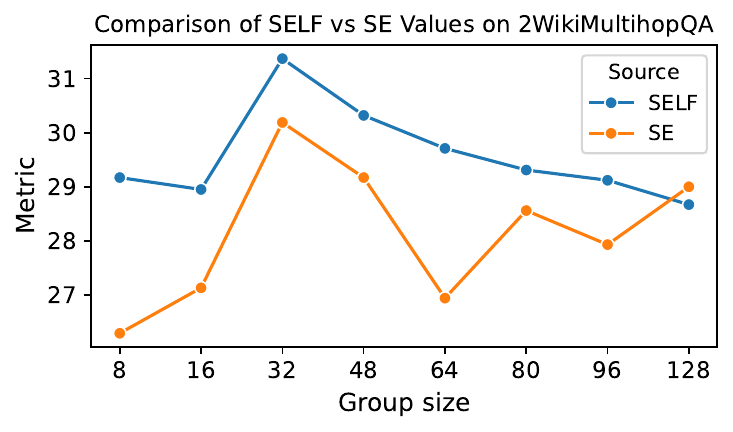}\vspace{-10pt}
    \caption{
    Compare the trade off with different group sizes of SE and SELF on 2WikiMultihopQA. The two grouping methods has the same neighbor window size $W=1024$.
    }
    \label{fig:trade_off}
    \vspace{-15pt}
\end{figure}
We see an improvement of summarizing tasks because having to go through smaller group sizes makes the models have a better understanding of the text. Moreover, our suspicion that SELF tends to perform better when the length is closer to the original context window is also confirmed since SELF performs better in tasks where the average context length is not significantly long such as MultiFieldQA, 2WikiMultihopQA, HotpotQA and TREC.

We can also observe the trend in trade off of accuracy in 2WikiMultihopQA task (see \cref{fig:trade_off}). Unlike SE, whose changes are more abrupt and unpredictable despite of the general trend is still decreasing (this can also be observed when SE doing other tasks with vary group sizes \cite{jin2024llmmaybelonglmselfextend}), SELF's accuracy seems to slowly decrease after reaching its peak.  In SE, group indices can be every different with different group sizes. In contrast, SELF's group sizes increase gradually, resulting in overlapping early groups. As the result, the differences between varying group sizes are relatively subtle compare to SE.

\subsubsection{LEval}

We tested one 4 different models using the same hyperparameter as in the LongBench v1 test (see the result \cref{tab:LEval}). 

\setlength{\tabcolsep}{4pt}
\begin{table*}[t]
\caption{Performance comparison of Llama2-7B, Llama2-13B, Qwen-7B and DeepSeek-R1-Distill-Qwen-7B before and after applying SE and SELF on LEval \cite{an2023levalinstitutingstandardizedevaluation}. The figure also includes performance of other fixed models on LEval. {\color{blue}*} means that the results are reported by LongLM \cite{jin2024llmmaybelonglmselfextend},  {\color{red}*} means that the results are run by us (single run). The best performance between the original, SE and SELF will be in \textbf{bold}.}\label{tab:LEval}\vspace{-10pt}
    \centering
    \begin{tabular}{l c c c c c c}
        \toprule
        \textbf{Model} & \textbf{Coursera} & \textbf{TOEFL} & \textbf{QuALITY} & \textbf{CodeU} & \textbf{SFiction} & \textbf{Avg.} \\
        \midrule
        \rowcolor{gray!20}
        Llama-2-7b-chat{\color{blue}*} & 29.21 & 51.67 & 37.62 & \textbf{1.11} & 60.15 & 35.95 \\
        \rowcolor{gray!20}
        SE-Llama-2-7b-chat{\color{blue}*} & 35.76 & 55.39 & \textbf{41.09}  & \textbf{1.11} & 57.81 & 38.23\\
        \rowcolor{gray!20}
        SELF-Llama-2-7b-chat{\color{red}*} & \textbf{36.19} & \textbf{56.88} & \textbf{41.09} & 0.00 & \textbf{60.94} & \textbf{39.02} \\
        \midrule
        \rowcolor{blue!20}
        Llama-2-13b-chat{\color{blue}*} & 35.75 & 60.96 & \textbf{42.57} & \textbf{1.11} & \textbf{60.15} & 40.11  \\ 
        \rowcolor{blue!20}
        SE-Llama-2-13b-chat{\color{blue}*} & \textbf{38.95} & \textbf{66.17} & 41.09 & \textbf{1.11} & 60.15 & \textbf{41.49} \\ 
        \rowcolor{blue!20}
        SELF-Llama-2-13b-chat{\color{red}*} & 37.93 & 64.31 & 39.11 & 0.00 & 57.03 & 39.68 \\ 
        \rowcolor{pink!20}
        Qwen-7B{\color{red}*} & 52.18 & 79.18 & 65.35 & 0.00 & \textbf{63.28} & 52.00 \\ 
        \rowcolor{pink!20}
        SE-Qwen-7B{\color{red}*} & 53.20 & 78.07 & 59.41 & 0.00 & 57.03 & 49.54  \\ 
        \rowcolor{pink!20}
        SELF-Qwen-7B{\color{red}*} & \textbf{53.34} & \textbf{80.67} & \textbf{66.83} & \textbf{4.44} & 62.5 & \textbf{53.56}  \\ 
        \midrule
        \textbf{\textit{Reasoning Model}} \\
        \midrule
        \rowcolor{cyan!20}
        DeepSeek-R1-Distill-Qwen-7B{\color{red}*} & \textbf{58.43} & \textbf{66.54} & \textbf{48.01} & 2.22 & 60.16 & \textbf{47.07} \\
        \rowcolor{cyan!20}
        SE-DeepSeek-R1-Distill-Qwen-7B{\color{red}*} & 54.21 & 66.17 & 40.59 & \textbf{6.66} & \textbf{62.4} & 45.81 \\
        \rowcolor{cyan!20}  
        SELF-DeepSeek-R1-Distill-Qwen-7B{\color{red}*} & 40.27 & 58.74 & 37.5 & 1.11 & 50.78 & 37.68 \\
        \midrule
        \textbf{\textit{Fixed Models}} \\
        \midrule
        Claude1.3-100k{\color{blue}*} & 60.03 & {83.64} & {73.76} & 17.77 & 72.65 & 65.97  \\
        GPT-4-32k & 75.58 & 84.38 & 82.17 & 25.55 & 74.99 & 73.11 \\
        Turbo-16k-0613{\color{blue}*} & 63.51 & 78.43 & 61.38 & 12.22 & 64.84 & 60.73 \\
        \midrule
        Chatglm2-6b-8k{\color{blue}*} & 43.75 & 53.90 & 40.59 & 2.22 & 54.68 & 34.69  \\
        XGen-7b-8k (2k-4k-8k){\color{blue}*} & 26.59 & 44.23 & 35.15 & 1.11 & 48.43 & 26.41 \\
        Chatglm2-6b-8k{\color{blue}*}  & 42.15 & 54.64 & 44.05 & 2.22 & 54.68 & 35.95 \\
        Chatglm2-6b-32k{\color{blue}*}  & 47.81 & 55.01 & 45.04 & 2.22 & 57.02 & 39.01 \\
        XGen-7b-8k{\color{blue}*} & 29.06 & 42.37 & 33.66 & 3.33 & 41.40 & 27.63 \\
        MPT-7b-65k{\color{blue}*} & 25.23 & 17.84 & 25.24 & 0.00 & 39.06 & 19.22 \\

        \bottomrule
    \end{tabular}
\end{table*}

\begin{itemize}[leftmargin=0.4cm, itemindent=.0cm, itemsep=0.0cm, topsep=0.1cm]
    \item \textbf{Llama-2-7B:} An improvement can be seen in every tasks except for CodeU. However, the difference in CodeU is not significant as the most correct models could answer only 1 correctly compare to 0 for model that apply SELF.
    \item \textbf{Llama-2-13B:} the SELF version seems to perform worse than the SE version in all tasks. We could not come up with the reason why there is such a difference despite Llama-2-7B and Llama-2-13B having the same architecture.
    \item \textbf{Qwen-7B:} There is a significant improvement compared to SE. There is an improvement compared to the raw model, but not significant.
    \item \textbf{Deepseek-R1-Distill-Qwen-7B\footnote{We modified the evaluation function for MCQ. After the reasoning process, models often start their conclusion with "Answer:", in which the original code \cite{an2023levalinstitutingstandardizedevaluation} will assume the answer to be "A" because "A" is the first capitalized letter}}: For this reasoning model, we forced the model to reason before giving answer by adding the open tag \verb|<think>|. An significant decrease in the accuracy of models after applying Self-Extend (both SE and SELF) can be observed. We suspect the Reinforcement Learning process to improve reasoning ability does have influence on this effect as reinforcement learning was applied on exact position and the model's reasoning ability wasn't optimized for group attention. We also witnessed models applying SE and SELF often stuck in reasoning loop which did not happen for the original model. However, this needs to be researched on more before making conclusion.
\end{itemize}

\section{Related Works}

\paragraph{Long context models.} Many recent large language models support extended context lengths, such as GPT-4~\cite{achiam2023gpt}, Claude, Qwen~\cite{yang2025qwen3}, LLaMA~\cite{grattafiori2024llama}, and Phi~\cite{abdin2024phi}. Notably, models like \texttt{Qwen2.5-7B-Instruct-1M}~\citep{qwen2.5-1m, qwen2.5} and Llama-3.1-Nemotron-8B-UltraLong-1M-Instruct~\citep{ulralong2025} are capable of handling context windows up to 1 million tokens, enabling long-range reasoning across extremely lengthy inputs. Multiple long context models exist optimized for long queries. LongChat~\citep{longchat2023,bai2024longbench} is a long context LLM designed for long context conversations. To train their model, LongChat is tested against their own long context testing suite and is trained with a context size of 32K. CLEX \cite{damonlpsg2023clex} is a long context LLM that works by using differential equations to scale positional embeddings to better support longer prompts. Code-Llama \cite{rozière2024codellamaopenfoundation} is a LLM model based on Llama 2 optimized for long context prompt performance. CodeLlama works by training the model on a longer context length of 16K.

\paragraph{Long context extension methods.} Most models increase the context length through fine-tuning, which still does not solve the problem of attention having a minimal affect at large relative distances. To solve this problem, other extension methods use a similar system where the position encodings modify the relative positions. Models with different context extension methods and their performance is mentioned in the above section above specific model performance. These include RoPE-based techniques such as Position Interpolation (PI)~\cite{chen2023extending}, NTK~\cite{peng2023ntk}, YaRN~\cite{peng2023yarn}, and SelfExtend~\cite{jin2024llm}; attention-architecture-based methods such as StreamingLLM~\cite{xiao2023efficient}, LM-Infinite~\cite{han2024lm}, LongLoRA~\cite{chen2023longlora}, Inf-LLM~\cite{xiao2024infllm}, and Landmark~\cite{mohtashami2023landmark}; as well as retrieval- and compression-based approaches such as Retrievers~\cite{xu2023retrieval}, LongLLMLingua~\cite{jiang2023longllmlingua}, and context compression~\cite{li2023compressing}.

\section{Conclusion}

We are successfully able to implement group attention with a custom function and apply it with a logistic growth function. From our analysis, we can conclude that SELF works better than SE when dealing with the same capacity, and SELF's behavior when increasing group size is more predictable.

Our logistic capacity model for grouping tokens yielded minor to major increases in most tests across LEval\footnote{With the exception of CodeU, a test where all methods performed poorly}. The method performed best on Llama-7b and Qwen-7B. On LongBench, our method performed better across most tests or saw only minor decreases in performance. By grouping tokens using a combined constant and logistic growth positional embedding layer, we allow the LLM to consider tokens at a far distance while keeping nearby tokens more relevant. SELF increases LLM prompt performance without sacrificing runtime performance nor modifying the prompt.

\clearpage
\section*{Limitations}
\begin{itemize}
    \item Although theoretically, SE and SELF has the basically same runtime complexity, SELF requires more complicated computations like $\ln$ instead of just \textsc{Floor}. As a result, running SELF takes a longer time than running SE.
    \item The grouping method was not tested in the variety of LLMs (only tested on LLama2-7B, Llama2-13B, Qwen-7B and Deepseek-R1-Distill-Qwen-7B) nor the variety of benchmarks.
    \item On reasoning models, running SELF the model would sometimes get stuck in a loop while thinking causing an unpredictable answer. This led to degraded performance compared to the raw model and SE.
    \item SELF still struggled on the CodeU benchmark compared to other models and would sometimes produce nonsensical outputs.
\end{itemize}

\bibliography{custom}

\end{document}